\documentclass[sigplan,nonacm,10pt,balance=false]{acmart}
\setcopyright{none}
\renewcommand\footnotetextcopyrightpermission[1]{}
\usepackage{amsmath,amssymb,booktabs,tabularx,array,xspace,graphicx}
\usepackage{placeins}
\usepackage{microtype}
\usepackage{enumitem}
\usepackage{xurl,pifont}
\hypersetup{colorlinks=true,linkcolor=black,citecolor=blue!45!black,urlcolor=blue!45!black}
\setlist[itemize]{leftmargin=*,topsep=3pt,itemsep=2pt,parsep=0pt}
\newcommand{\prjname}{\textsc{InfraBench}}

\newcolumntype{Y}{>{\raggedright\arraybackslash}X}
\definecolor{InfraBlue}{HTML}{235789}
\newcommand{\takeaway}[2]{\par\addvspace{4pt}\noindent\begingroup
\setlength{\fboxsep}{5pt}\colorbox{InfraBlue!7}{\parbox{\dimexpr\linewidth-2\fboxsep\relax}{\textbf{Finding #1.} #2}}\endgroup\par\addvspace{4pt}}
\newcommand{\iconClaudeCode}{\raisebox{-0.25ex}{\includegraphics[height=2ex]{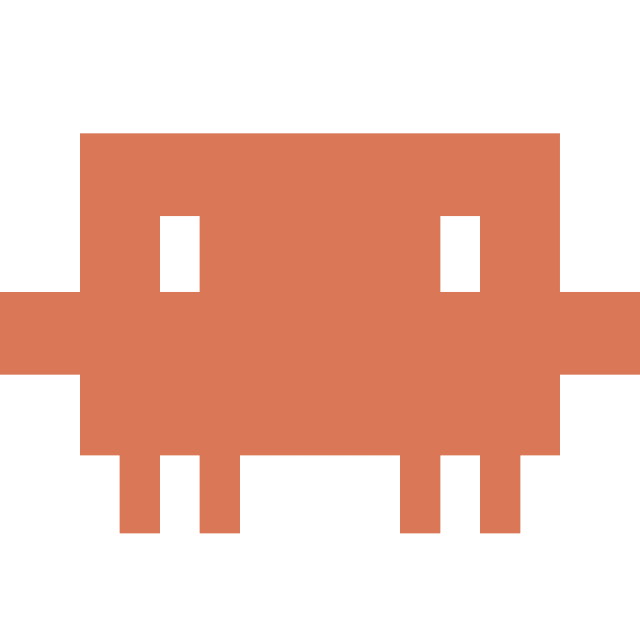}}}
\newcommand{\iconClaude}{\raisebox{-0.25ex}{\includegraphics[height=2ex]{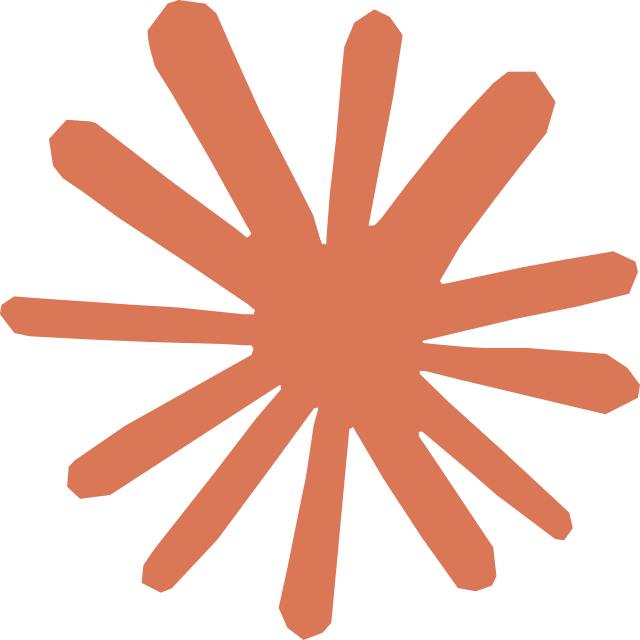}}}
\newcommand{\iconCursor}{\raisebox{-0.25ex}{\includegraphics[height=2ex]{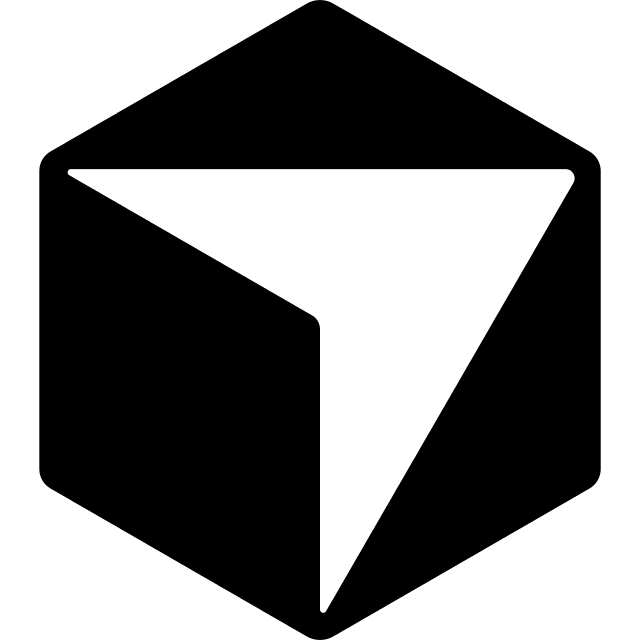}}}
\newcommand{\iconGrok}{\raisebox{-0.25ex}{\includegraphics[height=2ex]{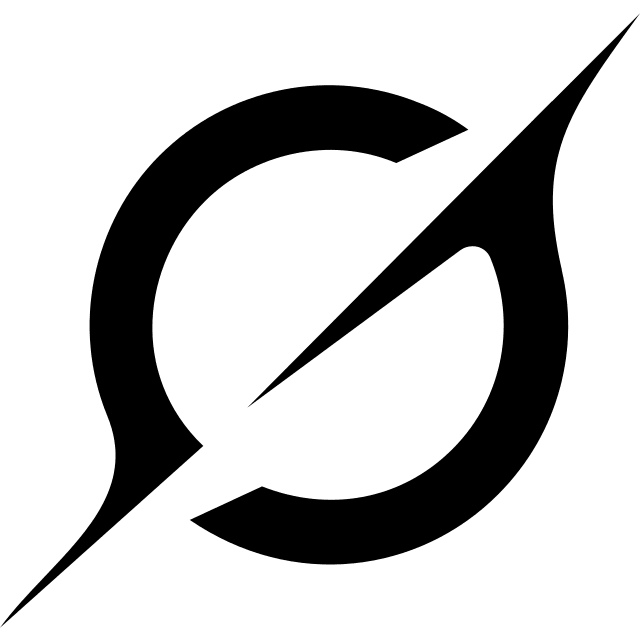}}}
\newcommand{\iconGeminiCLI}{\raisebox{-0.25ex}{\includegraphics[height=2ex]{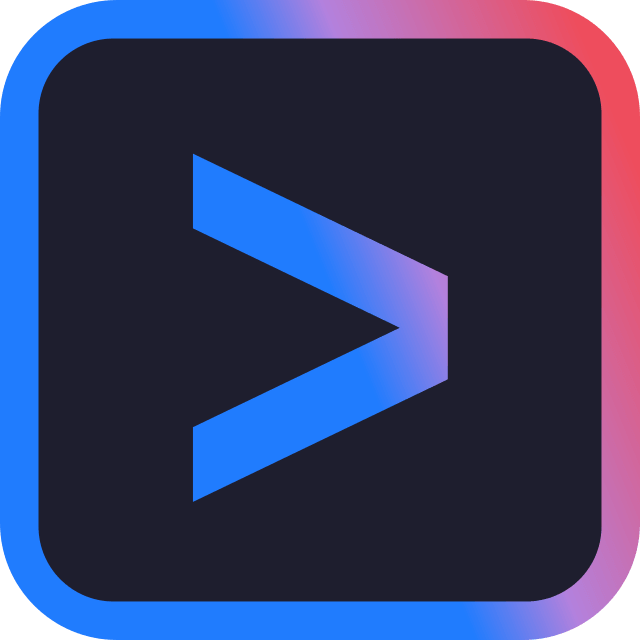}}}
\newcommand{\iconGemini}{\raisebox{-0.25ex}{\includegraphics[height=2ex]{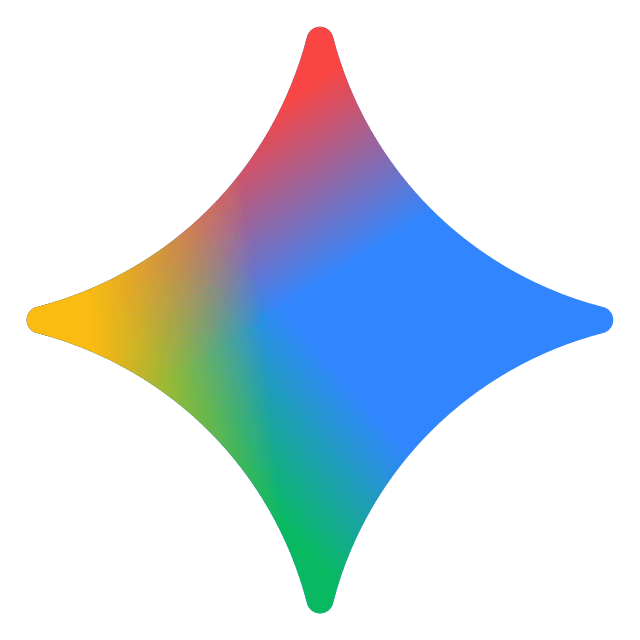}}}
\newcommand{\iconQoder}{\raisebox{-0.25ex}{\includegraphics[height=2ex]{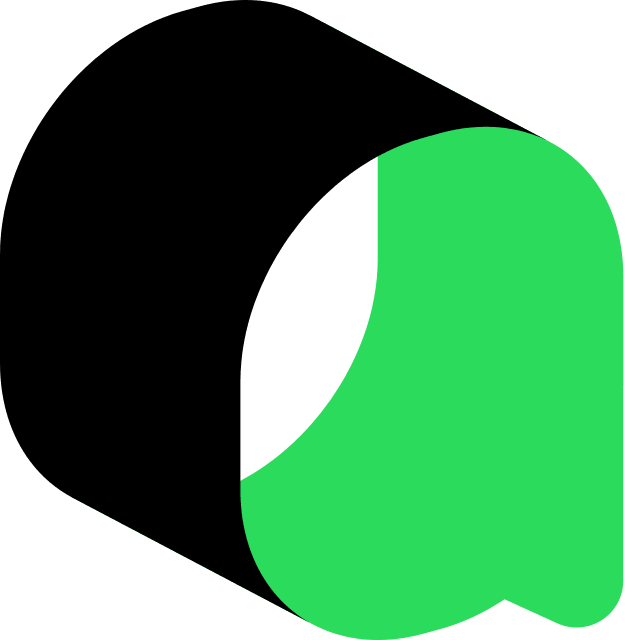}}}
\newcommand{\iconZhipu}{\raisebox{-0.25ex}{\includegraphics[height=2ex]{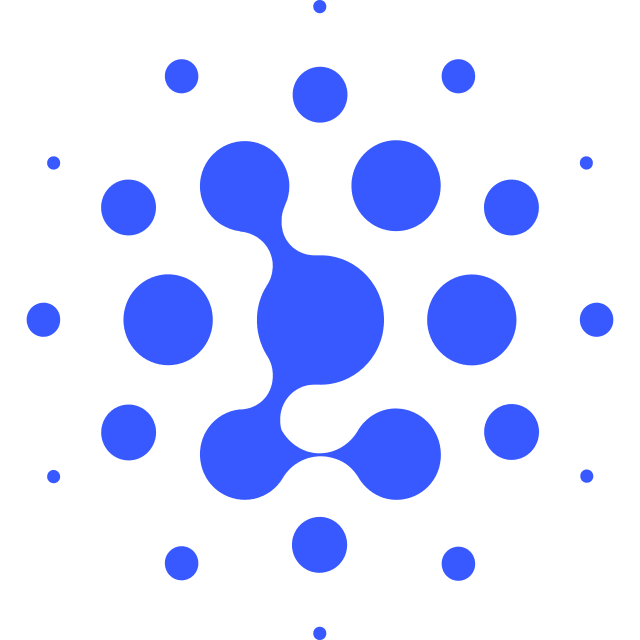}}}
\newcommand{\iconKimi}{\raisebox{-0.25ex}{\includegraphics[height=2ex]{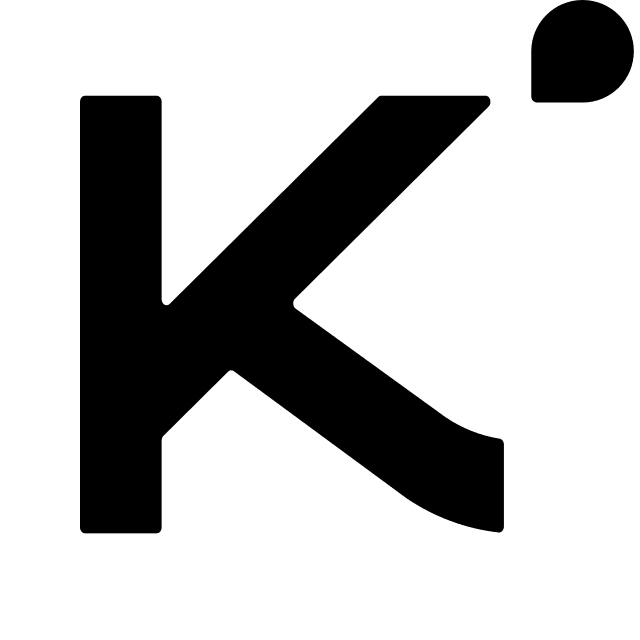}}}
\newcommand{\iconOpenCode}{\raisebox{-0.25ex}{\includegraphics[height=2ex]{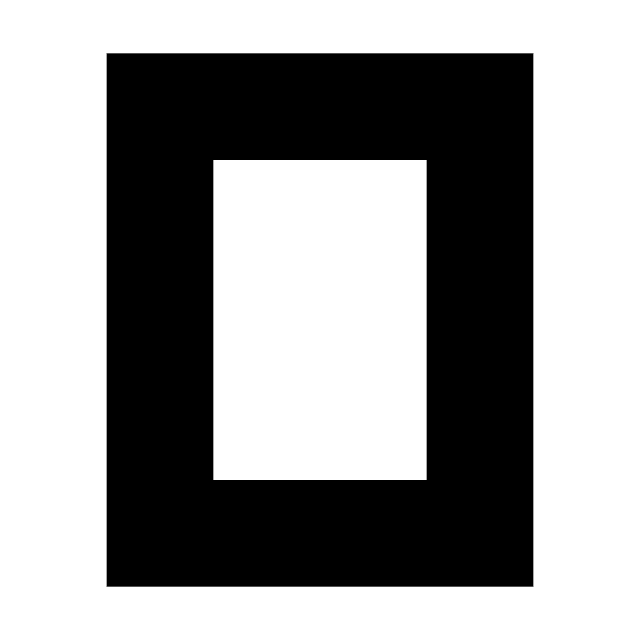}}}
\newcommand{\iconDeepSeek}{\raisebox{-0.25ex}{\includegraphics[height=2ex]{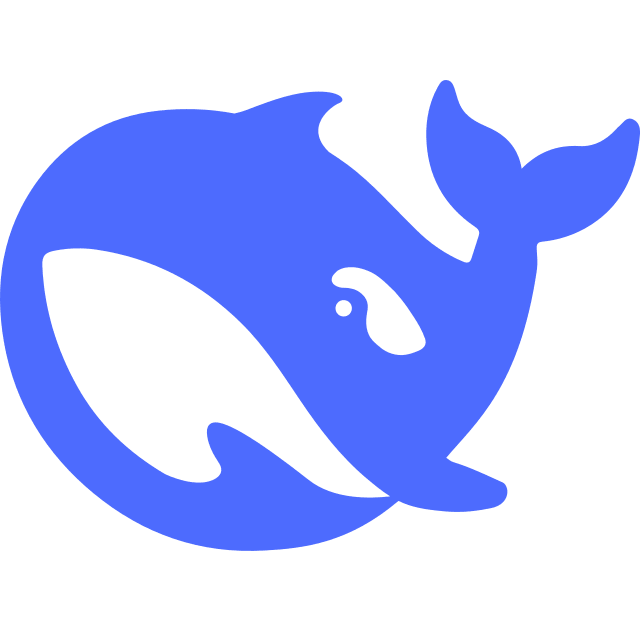}}}
\newcommand{\iconQwen}{\raisebox{-0.25ex}{\includegraphics[height=2ex]{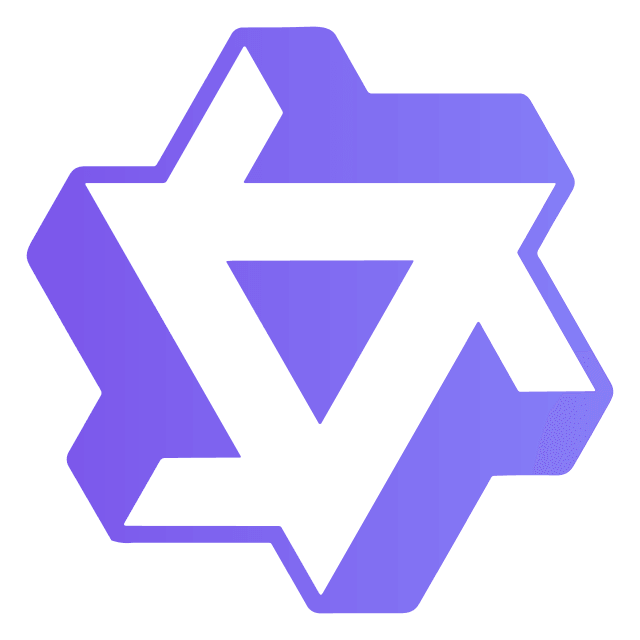}}}
\newcommand{\iconMiMo}{\raisebox{-0.25ex}{\includegraphics[height=2ex]{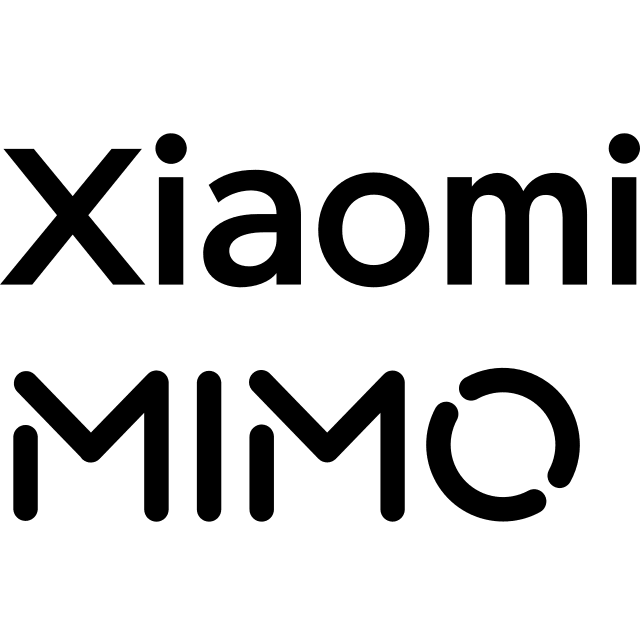}}}

\title{InfraBench: Evaluating Infrastructure Agents Across Layers, Lifecycle, and Risk}
\author{Yuan Gao}
\affiliation{\institution{University of Wisconsin--Madison}\city{Madison}\country{USA}}
\email{ygao355@wisc.edu}
\author{Zeren Yang}
\affiliation{\institution{University of Wisconsin--Madison}\city{Madison}\country{USA}}
\email{zyang667@wisc.edu}
\author{Junnan Li}
\affiliation{\institution{University of Wisconsin--Madison}\city{Madison}\country{USA}}
\email{jli2786@wisc.edu}
\author{Shawn (Wanxiang) Zhong}
\affiliation{\institution{University of Wisconsin--Madison}\city{Madison}\country{USA}}
\email{shawn.zhong@wisc.edu}
\author{Ahmed Dajani}
\affiliation{\institution{Iowa State University}\city{Ames}\country{USA}}
\email{adajani@iastate.edu}
\author{Mai Zheng}
\affiliation{\institution{Iowa State University}\city{Ames}\country{USA}}
\email{mai@iastate.edu}
\author{Andrea Arpaci-Dusseau}
\affiliation{\institution{University of Wisconsin--Madison}\city{Madison}\country{USA}}
\email{dusseau@cs.wisc.edu}
\author{Remzi Arpaci-Dusseau}
\affiliation{\institution{University of Wisconsin--Madison}\city{Madison}\country{USA}}
\email{remzi@cs.wisc.edu}
\renewcommand{\shortauthors}{Gao et al.}
\begin{document}
\begin{abstract}
Managing modern computing infrastructure has become a steadily harder problem due to the ever-increasing complexity. Recent advances in AI agents create a timely opportunity to automate infrastructure management tasks, but it remains unclear how well such agents can handle real-world infrastructure complexity.
We present \prjname{}, a benchmark for infrastructure agents organized around system layers, the operational lifecycle, and risk. The current prototype couples container, VM, and bare-metal backends with task-specific verifiers that distinguish functional repair from durability and cleanup, and analyzes retained action-trace findings for unsafe shortcuts.
We evaluate 18 agent--model configurations on 12 seed tasks, with three attempts per task. Mean effective scores range from 45.5\% to 89.2\%; the strongest configuration completes 77.8\% of attempts perfectly. In the available lifecycle checks, pass rates fall from 91.5\% for functional repair to 80.9\% for durability and 38.9\% for cleanup. These results show why immediate recovery is an incomplete measure of operational capability.
We use this evidence to motivate a planned expansion toward 40--60 tasks centered on managed cloud infrastructure and redesign after incidents, including provider-native Kubernetes networking. This extension is a research direction; the reported results remain those of the 12-task prototype.
\prjname{}, its tasks, evaluation harness, and live leaderboard are available at \href{https://infraben.ch}{infraben.ch}.

\end{abstract}
\maketitle
\pagestyle{plain}
\section{Introduction}
\label{sec:introduction}
As computing infrastructures continue to grow in scale and complexity, managing them has become a steadily harder problem. Even initial deployment now requires navigating diverse configurations across heterogeneous environments, from on-prem clusters to cloud interactions. Beyond deployment, continuous maintenance introduces additional burdens (e.g., upgrades and patching~\cite{hosek2013safe,maurer2015fail}, failure handling~\cite{du2017deeplog,han2022study,xu2019lessons,Om-FAST18-RFSCK,zheng2026fault}, migration and backup~\cite{clark2005live,hines2009post,mashtizadeh2011vmware,han2024revisiting}), all of which must be handled without disrupting service. This rising complexity turns infrastructure management into a persistent, long-standing challenge~\cite{lou2020partial,gunawi2014bugs,gunawi2016does,xu2019lessons}.

Recent advances in artificial intelligence (AI) agents~\cite{yao2023react,yang2024sweagent,shetty2024autonomous,chen2025aiopslab,ahmed2023recommending,jin2023assess} create a timely opportunity to revisit the challenge. A key question is whether AI agents can meaningfully automate these infrastructure-level tasks, and if so, to what extent they can handle the complexity and variability seen in the real world. Answering this question requires measuring the obligations an operator inherits after a repair. A service may become reachable while its configuration remains volatile, a peer loses data, or an incident record stays open. Each outcome looks successful under a narrow availability check, yet each leaves operational work unfinished.

\paragraph{From incident recovery to infrastructure redesign.}
Managed Kubernetes further couples cluster configuration with cloud resources such as subnets, network interfaces, and access permissions. An incident can expose a limitation in how these resources are organized. We use \emph{redesign after an incident} to describe tasks in which restoring the previous operating state is insufficient: the agent must change part of the architecture under explicit workload, availability, security, and resource constraints. It must select and implement a design, migrate the running system, and demonstrate that the triggering condition is handled. For example, recovering Pod creation after an address-allocation incident is a repair; establishing a valid allocation and rollout strategy that continues to work under the specified workload is a redesign obligation. This distinction changes both the task objective and the evidence needed to establish success.

Existing benchmarks address important parts of this problem. SREGym evaluates diagnosis and mitigation in live cloud-native stacks~\cite{clark2026sregym}; DevOps-Gym evaluates software-project workflows including building, monitoring, issue resolution, and testing~\cite{tang2026devopsgym}. Our proposed extension focuses on provider-native cloud behavior and explicitly required infrastructure redesign, validated through recurrence tests and lifecycle obligations. These are intended extensions of scope, not capabilities established by the current seed-task results (Table~\ref{tab:related-benchmarks-v2}).

\begin{table*}[t]
\centering\footnotesize
\setlength{\tabcolsep}{4pt}\renewcommand{\arraystretch}{1.06}
\begin{tabularx}{\textwidth}{@{}l Y Y Y@{}}
\toprule
\textbf{Benchmark} & \textbf{Task unit} & \textbf{Environment focus} & \textbf{Evidence of completion} \\
\midrule
SWE-bench~\cite{jimenez2024swebench} & Repository issue & Software repository & Tests of the resulting patch \\
Terminal-Bench~\cite{merrill2026terminalbench} & Terminal task & Containerized task environment & Task-specific state checks \\
DevOps-Gym~\cite{tang2026devopsgym} & Software DevOps workflow & Project build and runtime environment & Build, monitoring, fix, and test objectives \\
SREGym~\cite{clark2026sregym,sregymplatform2026} & Incident diagnosis and mitigation & Live stacks on Kubernetes; documented self-managed deployment & Diagnosis and mitigation oracles \\
\midrule
\prjname{} & Infrastructure operations and post-incident redesign & Containers, VM clusters, bare metal, and managed Kubernetes with provider-native networking & Recovery, design constraints, recurrence, migration, durability, cleanup, and action-risk evidence \\
\bottomrule
\end{tabularx}
\caption{\textbf{Evaluation targets and scope.} Comparison of task objectives, execution environments, and completion evidence.}
\label{tab:related-benchmarks-v2}
\end{table*}

\paragraph{The framework.}
We introduce \prjname{}, a benchmark suite that treats each task as a controlled infrastructure operation trial. Its design has four goals motivated by practical infrastructure management. These remain the basis for the cloud-focused extension:
\begin{itemize}
\item \textbf{Full-stack.} Operations span hardware, local system software, distributed systems, and user-facing applications. The backend must preserve the behavior under test, including hardware control and cross-node state.
\item \textbf{Full-lifecycle.} Success must account for deployment, live operation, maintenance, and decommissioning. A repair that works once may fail under load or after a restart.
\item \textbf{Risk-aware.} An agent can satisfy its immediate objective while expanding the blast radius~\cite{gunawi2016does,han2022study}. Actions and preservation invariants must therefore be examined alongside completion.
\item \textbf{Realistic and extensible.} Tasks should reflect operational incidents and constraints, execute on appropriate container, VM, or bare-metal backends, and support contributions from practitioners.
\end{itemize}

Four complementary sources inform the tasks: interviews with infrastructure providers and practitioners at three university centers~\cite{uw-doit,uw-chtc,uw-cdis} and a cross-city testbed~\cite{islam2025design}; open-source repositories and issue trackers, including Slurm~\cite{yoo2003slurm}, Pelican~\cite{pelicanplatform}, and Ceph~\cite{weil2006ceph}; commercial cloud documentation; and systems-research prototypes. These sources capture both documented failures and operational constraints that are seldom explicit in benchmark prompts. The resulting workflow separates agent-visible instructions from hidden evaluation context, selects a faithful backend, and evaluates the state left after the agent finishes (\S\ref{sec:design}).

\paragraph{Current evidence and intended expansion.}
The implemented prototype contains 12 seed tasks across four infrastructure layers; these experiments do not include managed EKS/AKS or redesign tasks. We evaluate 18 configurations spanning five agent CLIs, with three attempts per task (\S\ref{sec:setup}). We plan to expand toward 40--60 tasks with an emphasis on cloud operations and redesign; this is a construction target, separate from the results reported here. The current study yields three main findings:
\begin{itemize}
\item \textbf{High average scores leave a reliability gap.} Mean effective scores range from 45.5\% to 89.2\%, while the strongest configuration achieves perfect completion on 77.8\% of attempts (\S\ref{sec:exp-leaderboard}).
\item \textbf{Failures persist after functional recovery.} Functional checks pass at 91.5\%, durability checks at 80.9\%, and cleanup checks at 38.9\%. Missing cleanup and deployment residue occur across all 18 configurations (\S\ref{sec:exp-lifecycle}--\ref{sec:exp-modes}).
\item \textbf{Completion, risk, and cost expose different trade-offs.} The audit flags 27 of 16{,}511 recorded commands, including evaluator probing and destructive shortcuts. Estimated campaign costs vary by more than two orders of magnitude; the most expensive configurations are not the most reliable (\S\ref{sec:exp-risk}--\ref{sec:exp-cost}).
\end{itemize}

\paragraph{Contributions and availability.}
We contribute a framework that couples layer-appropriate execution with lifecycle and risk evaluation; a practitioner-informed seed suite with check-level scoring; and an empirical study of reliability, failure modes, and operating cost. Building on these implemented components, we specify a direction for cloud-native repair and infrastructure redesign (\S\ref{sec:design-extension}). We release the tasks, harness, and leaderboard at \href{https://infraben.ch}{infraben.ch} to support broader evaluation and task contributions.
A preliminary version appeared at HotInfra~'26~\cite{gao2026beyond} and reported seven configurations. This full paper evaluates 18 configurations and adds repeated-attempt reliability, lifecycle analysis, an action-trace risk audit, and cost analysis. All reported experimental results refer to the current campaign.

\section{\prjname{} Design \& Implementation}
\label{sec:design}
\begin{figure*}[t]
\centering
\includegraphics[width=\textwidth]{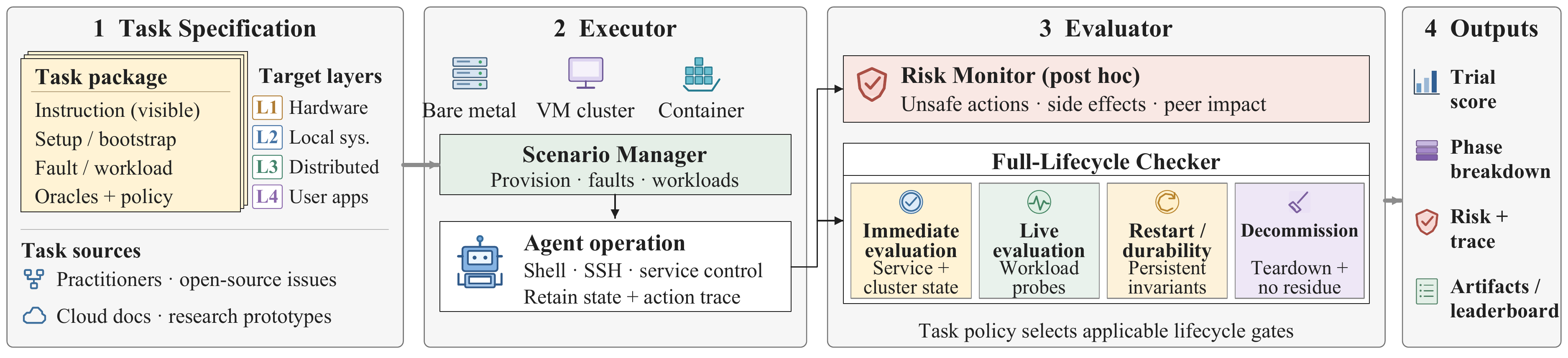}
\Description{Four grouped stages connected by arrows: task specification with layer annotations, executor with scenario preparation and agent operation, evaluator with parallel risk and lifecycle paths, and trial outputs.}
\caption{\textbf{\prjname{} workflow.} Task packages combine an agent-visible objective with hidden setup, fault, oracle, and lifecycle definitions. Layer-aware execution produces infrastructure state and an action trace. The evaluator checks state across policy-selected lifecycle gates and audits actions for risk, producing scores and supporting evidence.}
\label{fig:design-overview}
\end{figure*}

\prjname{} evaluates a controlled infrastructure operation trial through four components: \emph{Task Specification}, \emph{Executor}, \emph{Evaluator}, and \emph{Output} (Figure~\ref{fig:design-overview}). An agent receives an operational objective and acts through the interfaces available to a human operator. The evaluator then checks the resulting infrastructure state and audits the recorded actions.

\paragraph{Infrastructure layers.}
Tasks are annotated with four layers. \textbf{L1 Hardware} covers physical control, including BMC/IPMI and power recovery. \textbf{L2 Local Systems} covers host software such as the OS, local networking, and container runtimes. \textbf{L3 Distributed Systems} covers networked systems whose behavior depends on coordination or cross-node state, such as Ceph~\cite{weil2006ceph}, Slurm~\cite{yoo2003slurm}, and Fabric~\cite{androulaki2018hyperledger}. \textbf{L4 User Applications} covers user-facing services running on local or distributed infrastructure. Layer annotations guide backend selection and the invariants a verifier must inspect; they do not require every task to exercise every layer.

\subsection{Task Specification}
Each task package separates two kinds of information. The \emph{agent-visible instruction} describes the requested operation. The \emph{hidden evaluation context} defines setup and bootstrap requirements, workloads and injected faults, verification oracles, and the lifecycle policy that selects the applicable gates. This separation prevents task-specific grading instructions from becoming part of the agent's prompt. Task authors can add infrastructure-specific constraints, including requirements elicited from university-center operators, without changing the execution workflow.

\subsection{Executor}
The executor instantiates the task on a backend that preserves the operation's relevant semantics. Its \emph{Backend Selector} maps the task to containers, VM clusters, or bare-metal hosts with out-of-band control. For example, a container cannot substitute for a physical power-control task, and a single process cannot reproduce a failure that depends on peer state.

The \emph{Scenario Manager} provisions the initial state, applies resource limits and fault conditions, and opens the agent operation window. The agent uses shell, SSH, service-management, and task-appropriate control interfaces. Verifiers and oracle scripts remain outside the agent-visible task specification. After the window closes, the resulting state and recorded trajectory are retained for evaluation. Each new attempt starts from a freshly provisioned state (\S\ref{sec:setup-protocol}).

\subsection{Evaluator}
The evaluator measures both operational correctness and risk. These signals answer different questions: whether the resulting infrastructure satisfies its obligations, and whether the agent took a hazardous route to that result.

\paragraph{Full-Lifecycle Checker.}
The checker separates short-term recovery from longer-lived correctness through four policy-selected gates:
\begin{itemize}
\item \textbf{Immediate evaluation} compares the baseline and post-operation state to determine whether the requested objective was met.
\item \textbf{Live evaluation} uses sustained workloads or periodic probes to test continued availability and detect drift or delayed degradation.
\item \textbf{Restart/durability} restarts the relevant service or resource and tests persistent configuration and post-restart invariants.
\item \textbf{Decommission} checks that requested resources are removed or the infrastructure is restored without leaks or residual incident state.
\end{itemize}
Not every task implements every gate. In the current campaign, we analyze scored checks in three derived categories (Functional, Durability, and Cleanup) according to the obligation each check tests (\S\ref{sec:exp-lifecycle}). These categories summarize the available checks; they are not a sequential survival experiment over the same population.

\paragraph{Risk Monitor.}
Operational risk and side effects are first-class signals alongside lifecycle correctness. In the current implementation, candidate actions from archived trajectories are reviewed by an LLM judge against a fixed seven-type danger taxonomy (\S\ref{sec:exp-risk}). It examines commands in context and emits structured findings, including destructive operations, broad safety bypasses, and evaluator-harness probing. This is a post hoc audit of retained logs; it does not block commands during execution.

Verifier preservation checks complement the action audit by testing observable consequences, such as whether baseline data remain intact, a peer remains reachable, or a repair survives a restart. An action flag is review evidence, whereas a failed preservation check establishes a violated invariant. Neither signal alone measures the full severity of a production incident.

\subsection{Metrics}
\label{sec:metrics}
A task-specific verifier returns reward $R\in[0,1]$. For verifiers with scored checks, $R$ is the weighted fraction passed. Check weights are derived once from the campaign and frozen; the two aggregate-only graders remain binary (Appendix~\ref{app:weighting}). We report four metrics, assigning zero to missing attempt slots:

\paragraph{Mean effective score.}
We average the three attempt rewards for each task, then average the 12 task means and express the result as a percentage. Each task therefore has equal weight in the headline score, regardless of its number of verifier checks.

\paragraph{Attempt Pass@$\tau$.}
This is the fraction of the 36 scheduled attempts with $R\geq\tau$. We use $\tau=1$ for perfect completion and $\tau=0.5$ for substantial completion. These are empirical threshold pass rates, distinct from the usual code-generation pass@$k$ estimator.

\paragraph{Best-of-N@$\tau$.}
With $N=3$, this is the fraction of the 12 tasks whose best attempt reaches $R\geq\tau$. It describes the observed benefit of selecting the best of three results; it does not estimate unattended deployment reliability.

\paragraph{Mean $\pm$ SEM.}
We compute SEM over the 12 per-task means: $\mathrm{SEM}=s/\sqrt{12}$, where $s$ is their sample standard deviation in percentage points. It describes variation across this task set, rather than uncertainty over a representative population of infrastructure incidents. Appendix~\ref{app:metrics} gives the formulas.

\subsection{Output}
For each trial, \prjname{} retains the score, check and phase breakdowns, risk findings, execution timeline, and trajectory artifacts. Leaderboard summaries aggregate these records while preserving the evidence needed to attribute a failure to an unmet lifecycle obligation or side effect. This makes a partial repair distinguishable from an operationally complete one.

\subsection{Planned Extension: Cloud Repair and Redesign}
\label{sec:design-extension}
The next task set will focus on cloud infrastructure, including managed Kubernetes environments whose relevant behavior depends on provider services. The current prototype and Figure~\ref{fig:design-overview} establish the execution and evaluation structure; the following task profiles and validation requirements are planned rather than implemented results.

\paragraph{Provider-native environments.}
Target profiles include Amazon EKS with Amazon VPC CNI~\cite{awsvpccni2026} and Azure AKS with a specified Azure CNI configuration~\cite{azurecni2026}. A cloud name alone is not an environment specification. Each profile must pin the Kubernetes and networking versions, address-management mode, permissions, topology, and workload assumptions relevant to the task. For example, Pod-address allocation can depend on subnet capacity and node-networking resources, while a proposed change must be permitted by the agent's cloud access policy. Diagnosis therefore combines Kubernetes observations with the relevant provider resources and permissions. Self-managed clusters remain useful for controlled experiments, while provider-specific claims require runs against the corresponding provider profile.

\paragraph{A distinct redesign objective.}
A redesign task starts with an incident or on-call escalation and states the operational requirement that the existing design fails to meet. The agent must diagnose the limitation, choose and implement a permitted change, and migrate the running system. The objective specifies required behavior and constraints rather than a reference patch or a prescribed sequence of commands. Admissible changes may involve network or address-allocation design, workload placement, isolation boundaries, or resilience configuration, according to the task.

\paragraph{Example: a recurring network-capacity incident.}
A proposed task begins when a workload increase exhausts the configured Pod-address capacity. Restoring a few Pods does not satisfy the redesign objective. The agent must establish a provider-compatible capacity and rollout design within explicit resource and availability limits, preserve required connectivity and isolation during migration, and show that the specified workload can be sustained. Validation would replay the triggering workload and test rollout or node replacement against the revised system. The permitted design depends on the pinned cloud profile; the task does not assume that an EKS repair transfers unchanged to AKS. This is an illustrative task specification, not a reported experiment.

\paragraph{Verification beyond service restoration.}
The proposed verifier separately checks immediate recovery, satisfaction of the new design constraints, resistance to the stated recurrence trigger, and preservation during migration. Applicable lifecycle gates then test persistence, rollback, and cleanup. Merely suppressing an alert, relaxing isolation, or deleting the state under repair cannot satisfy these obligations. Different valid architectures can pass the same behavioral tests; reproducing one reference configuration is not the evaluation goal.

\section{Experimental Setup}
\label{sec:setup}

\subsection{Tasks and Testbed}
\label{sec:setup-tasks}
The results in this paper use the current 12-task seed suite; the planned managed-cloud and redesign extension in \S\ref{sec:design-extension} is not included. The prototype tasks span the four infrastructure layers (Table~\ref{tab:task-taxonomy}; full catalog with difficulty and check counts in Appendix~\ref{app:tasks}), drawn from the four sources described in \S\ref{sec:introduction}: practitioner interviews and incident accounts, open-source issue trackers, cloud platform documentation, and systems-research prototypes. Each task specifies a target infrastructure layer, a fault or drift condition, and a lifecycle policy that determines which of the four Evaluator gates (\S\ref{sec:design}) apply. The set deliberately mixes recovery (power, crash, hung-node, RAID, WAL), deployment (Ceph bootstrap), and drift-repair (scheduler, connection-pool, federation-identity) scenarios, so no single operation type dominates.

Depending on its layer, a task is provisioned as a Docker container (L4), a multi-node VM cluster over libvirt/KVM (L2--L4), or a bare-metal cluster with out-of-band IPMI/BMC control (L1, L3). All trials run on the CloudLab Wisconsin testbed~\cite{cloudlab} on c220g1 nodes (two 8-core Xeon E5-2630~v3, 128\,GB RAM, dual 10\,GbE), so every agent operates against the same physical hardware class regardless of backend. Each attempt starts from a freshly provisioned environment: the Scenario Manager (\S\ref{sec:design}) rebuilds the target state---including injected faults---before the agent operation window opens, so consecutive attempts of the same task are independent. During the window the agent holds root privileges and the same operational interfaces a human operator would use (shell, SSH to peer nodes, service managers, and, for L1 tasks, the IPMI control plane); verifier and oracle scripts are never exposed to the agent.

\begin{table*}[t]
\centering\footnotesize
\setlength{\tabcolsep}{5pt}\renewcommand{\arraystretch}{1.05}
\begin{tabularx}{\textwidth}{@{}l c l Y@{}}
\toprule
\textbf{Task} & \textbf{Layer} & \textbf{Backend} & \textbf{Operation or fault} \\
\midrule
IPMI Power Recovery & L1 & Bare-metal cluster & Restore power through out-of-band control \\
Cassandra NIC Split-Brain & L2 & Bare-metal cluster & Repair local network-stack misconfiguration \\
Cassandra Dead Node Removal & L3 & Bare-metal cluster & Remove a crashed node while preserving the cluster \\
Cassandra Hung Recovery & L3 & Bare-metal cluster & Recover a hung node \\
Cassandra CORDS Propagation & L3 & VM cluster & Repair replica inconsistency; adapted from CORDS~\cite{ganesan2017redundancy} \\
Ceph Bootstrap & L3 & VM cluster & Deploy a distributed storage system \\
Ceph Pool Degraded & L3 & VM cluster & Recover degraded distributed resources \\
Fileserver RAID10 & L3 & VM cluster & Repair storage degradation in a distributed service \\
Slurm/Puppet Cascade & L3 & VM cluster & Reconcile scheduler configuration drift \\
DB WAL Recovery & L4 & Container & Recover database state; adapted from Terminal-Bench~\cite{merrill2026terminalbench} \\
Postgres/PgBouncer Drift & L4 & VM cluster & Repair application and connection-pool drift \\
Pelican Key Mismatch & L4 & VM cluster & Reconcile federation identity and close incident state \\
\bottomrule
\end{tabularx}
\caption{\textbf{Current task set.} Twelve seed tasks across four infrastructure layers. Appendix~\ref{app:tasks} gives difficulty labels and check counts.}
\label{tab:task-taxonomy}
\end{table*}

\subsection{Agents and Models}
\label{sec:setup-agents}
We evaluate 18 agent--model configurations spanning five coding-agent command-line interfaces: Claude Code, Cursor CLI, Gemini CLI, OpenCode, and Qoder CLI, paired with models from nine vendors (Anthropic, Google, xAI, Cursor, DeepSeek, Xiaomi, Zhipu, Moonshot, and Alibaba; Table~\ref{tab:agent-success}). A configuration couples an agent CLI---which supplies the scaffolding: the tool-use loop, context management, and shell integration---with an underlying model that does the reasoning; the two are not independent, and \S\ref{sec:exp-leaderboard} shows a range of more than 25 points within the same CLI as the underlying model changes. Every configuration pins a fixed model checkpoint: vendor model routers, whose backend selection changes without notice, are excluded so that a reported result stays reproducible against a named model version.

All CLIs run with vendor-default settings and no system-prompt customization beyond the task instruction, and no human intervenes during the operation window. The CLI defaults---context management, tool-call policy, and any built-in system prompt---are part of the configuration under test rather than a nuisance variable: a configuration is the pair (CLI version, model checkpoint), and both are pinned and recorded, so the comparison across models within one CLI holds the scaffolding fixed. Each attempt is bounded by a per-task agent time budget (typically 30 minutes), after which the environment is frozen and handed to the verifier; the exact CLI version used by each trial is recorded in its trajectory artifact. Each configuration is given the same agent-visible instruction and the same operation window; the hidden evaluation context (target layer, fault conditions, oracle scripts, lifecycle policy) is identical across configurations for a given task, to reduce avoidable environmental differences between configurations.

\subsection{Evaluation Protocol}
\label{sec:setup-protocol}
Each configuration is allocated three attempts on every task, each attempt on a freshly provisioned environment, scored by the task's verifier under the difficulty-weighted rubric of \S\ref{sec:metrics} (Appendix~\ref{app:weighting}). To distinguish occasional success from repeated completion, we report Attempt Pass@$\tau$ and Best-of-N@$\tau$ (\S\ref{sec:metrics}) alongside the mean: together they separate configurations that solve a task reliably from those that solve it once.

Failures in provisioning, environment start, or agent setup occur before the operation window. They are retried without a score so that setup noise does not penalize an agent. Once the window opens the attempt is scored: a timeout or an abnormal exit still runs the verifier, and damage the agent itself causes---an unreachable peer, a broken route, a disabled interface---is graded by the checks it fails rather than excused as testbed noise, so an agent cannot earn a retry by breaking its own environment. The residual case is an environment left so damaged that the verifier cannot run at all; such an attempt is reported as uncovered and receives zero in the effective-score denominator, rather than being silently retried. Every configuration is evaluated on the same 12 tasks under the same protocol.

\section{Results}
\label{sec:experiment}
We organize the study around four questions: how reliably configurations complete a task; which operational obligations remain unmet; what risk is visible in their actions; and how reliability relates to cost. All results use the campaign and scoring protocol in \S\ref{sec:setup}.

\subsection{Overall Performance and Reliability}
\label{sec:exp-leaderboard}
Table~\ref{tab:agent-success} reports the 18 agent--model configurations. Mean effective scores range from 45.5\% to 89.2\%. Fable~5 with Claude Code has the highest mean, followed by Grok~4.5 with Cursor CLI at 84.3\%. These are configuration comparisons: the CLI's tool loop and context management are part of the system under test.

\begin{table*}[t]
\centering
\footnotesize
\setlength{\tabcolsep}{4pt}
\renewcommand{\arraystretch}{1.07}
\begin{tabular*}{\textwidth}{@{\extracolsep{\fill}} r l l r r r r @{}}
\toprule
\textbf{\#} & \textbf{Agent} & \textbf{Model} & \textbf{Mean} & \textbf{Pass@1} & \textbf{Pass@0.5} & \textbf{Best-of-N@1} \\
\midrule
1  & \iconClaudeCode~Claude Code & \iconClaude~Fable 5                & \textbf{89.2} $\pm$ 5.6  & 77.8\% (28) & 91.7\% (33) & 83.3\% (10) \\
2  & \iconCursor~Cursor CLI            & \iconGrok~Grok 4.5                       & 84.3 $\pm$ 7.9          & 75.0\% (27) & 86.1\% (31) & 83.3\% (10) \\
3  & \iconClaudeCode~Claude Code & \iconClaude~Claude Opus 5          & 83.1 $\pm$ 7.5          & 72.2\% (26) & 86.1\% (31) & 75.0\% (9) \\
4  & \iconQoder~Qoder CLI        & \iconQwen~Qwen3.8 Max              & 81.4 $\pm$ 8.6          & 69.4\% (25) & 77.8\% (28) & 75.0\% (9) \\
5  & \iconClaudeCode~Claude Code & \iconClaude~Claude Opus 4.8        & 80.2 $\pm$ 7.5          & 63.9\% (23) & 86.1\% (31) & 75.0\% (9) \\
6  & \iconQoder~Qoder CLI        & \iconKimi~Kimi K3                  & 77.3 $\pm$ 8.4          & 66.7\% (24) & 77.8\% (28) & 75.0\% (9) \\
7  & \iconClaudeCode~Claude Code & \iconClaude~Claude Sonnet 5        & 77.2 $\pm$ 7.4          & 63.9\% (23) & 80.6\% (29) & 75.0\% (9) \\
8  & \iconGeminiCLI~Gemini CLI   & \iconGemini~Gemini 3.6 Flash       & 75.5 $\pm$ 9.5          & 66.7\% (24) & 72.2\% (26) & 75.0\% (9) \\
9  & \iconQoder~Qoder CLI        & \iconZhipu~GLM 5.2                 & 74.0 $\pm$ 8.2          & 61.1\% (22) & 77.8\% (28) & 75.0\% (9) \\
10 & \iconCursor~Cursor CLI            & \iconCursor~Composer 2.5                 & 72.6 $\pm$ 9.3          & 58.3\% (21) & 75.0\% (27) & 66.7\% (8) \\
11 & \iconGeminiCLI~Gemini CLI   & \iconGemini~Gemini 3.5 Flash       & 72.1 $\pm$ 11.0         & 63.9\% (23) & 66.7\% (24) & 66.7\% (8) \\
12 & \iconGeminiCLI~Gemini CLI   & \iconGemini~Gemini 3.1 Pro Preview & 70.9 $\pm$ 8.0          & 52.8\% (19) & 72.2\% (26) & 58.3\% (7) \\
13 & \iconQoder~Qoder CLI        & \iconKimi~Kimi K2.7 Code           & 68.9 $\pm$ 8.8          & 55.6\% (20) & 72.2\% (26) & 75.0\% (9) \\
14 & \iconOpenCode~OpenCode      & \iconDeepSeek~DeepSeek V4 Flash    & 60.6 $\pm$ 10.3         & 47.2\% (17) & 63.9\% (23) & 66.7\% (8) \\
15 & \iconOpenCode~OpenCode      & \iconMiMo~MiMo V2.5          & 53.0 $\pm$ 10.6         & 38.9\% (14) & 61.1\% (22) & 50.0\% (6) \\
16 & \iconClaudeCode~Claude Code & \iconClaude~Claude Sonnet 4.6      & 52.6 $\pm$ 10.1         & 36.1\% (13) & 52.8\% (19) & 50.0\% (6) \\
17 & \iconGeminiCLI~Gemini CLI   & \iconGemini~Gemini 3.1 Flash-Lite  & 46.1 $\pm$ 11.2         & 33.3\% (12) & 50.0\% (18) & 41.7\% (5) \\
18 & \iconOpenCode~OpenCode      & \iconDeepSeek~DeepSeek V4 Pro      & 45.5 $\pm$ 11.4         & 36.1\% (13) & 52.8\% (19) & 50.0\% (6) \\
\bottomrule
\end{tabular*}
\caption{\textbf{Overall results: partial credit and repeated-attempt reliability.} Mean effective score $\pm$ SEM over 12 per-task means, in percentage points. Attempt Pass@$\tau$ counts attempts with $R\geq\tau$; parenthesized numerators are out of 36 scheduled attempts. Best-of-N@1 counts tasks with at least one perfect attempt out of three; its numerators are out of 12 tasks. Missing slots receive zero. Checkpoint names are those recorded in the campaign.}
\label{tab:agent-success}
\end{table*}
\paragraph{Average reward overstates perfect completion.}
Fable~5 reaches a perfect score in 28 of 36 attempts (77.8\%), despite its 89.2\% mean reward. Grok~4.5 scores 84.3\% on average but completes 27 of 36 attempts perfectly (75.0\%). Best-of-three selection raises perfect task coverage to 10 of 12 tasks for both configurations, leaving two tasks without a perfect result even after three attempts. Partial credit and retry success therefore answer different operational questions from single-attempt reliability.

\paragraph{The underlying model matters within a CLI.}
Claude Code spans 52.6\% with Sonnet~4.6 to 89.2\% with Fable~5, a range of more than 25 points. Gemini CLI spans 46.1\% to 75.5\%, and Qoder CLI spans 68.9\% to 81.4\%. These within-CLI comparisons hold the scaffolding family fixed, but do not establish a model-only ranking across different CLIs.
\takeaway{1}{The best mean score is 89.2\%, but the best observed perfect-completion rate is 77.8\%. Reporting both prevents partial repairs from being mistaken for dependable completion.}

\subsection{Which Tasks Separate Configurations?}
\label{sec:exp-perproblem}
\begin{figure*}[t]
\centering
\includegraphics[width=\textwidth]{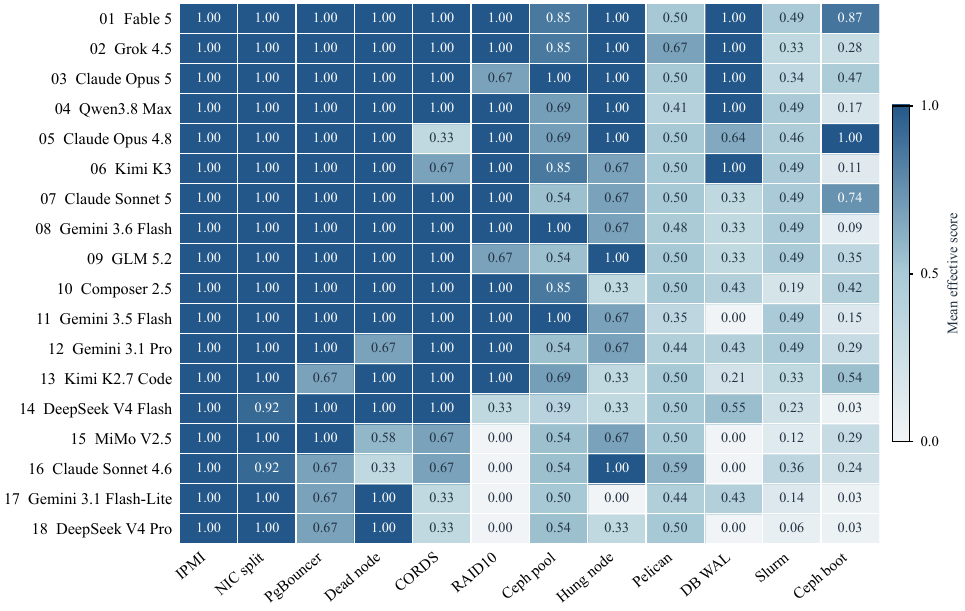}
\Description{Eighteen rows of model configurations and twelve task columns, with three-attempt mean scores shown in every cell.}
\caption{\textbf{Per-task effective scores.} Configurations follow leaderboard order; tasks are ordered by mean score, easiest at left. Each cell gives the three-attempt effective-score mean, rounded to two decimals. A sequential scale runs from low reward (light) to high reward (dark); numbers preserve partial-credit differences without relying on color. Appendix~\ref{app:perproblem} provides the tabular values.}
\label{fig:per-problem}
\end{figure*}

Figure~\ref{fig:per-problem} decomposes the headline score by task. Hardware power recovery and NIC split-brain repair are solved by nearly every configuration. The hardest tasks expose different limits: only Opus~4.8 averages a perfect score on Ceph Bootstrap, while five configurations average a perfect score on DB WAL Recovery. RAID10 recovery and Cassandra Hung Recovery span the full $[0,1]$ range across configurations, making them particularly discriminative in this task set.

These results also bound layer-level conclusions. The single L1 task shows that the evaluated agents can use its out-of-band power controls; it does not establish broad hardware-management competence. Seven of the 12 tasks are assigned to L3, so the prevalence of distributed-system failures partly reflects the task mix.

\subsection{Which Obligations Remain After Repair?}
\label{sec:exp-lifecycle}
We group the scored checks available in the lifecycle analysis into \emph{Functional}, \emph{Durability}, and \emph{Cleanup}, using a deterministic rule over verifier check names (Appendix~\ref{app:lifecycle-mapping}). These are check-level summaries with different task mixes and denominators, rather than consecutive stages traversed by the same trials.

\begin{figure}[t]
\centering
\includegraphics[width=\columnwidth]{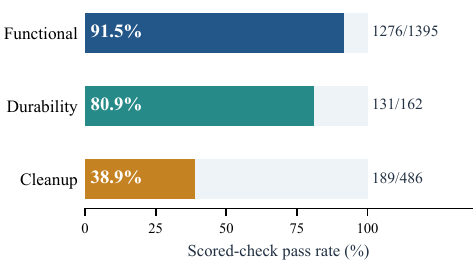}
\Description{Independent functional, durability, and cleanup check pass rates: 91.5, 80.9, and 38.9 percent.}
\caption{\textbf{Check pass rates by obligation.} Separate check populations, with passed/total counts; categories are not sequential stages.}
\label{fig:lifecycle}
\end{figure}

Functional checks pass at 91.5\% (1276/1395), durability checks at 80.9\% (131/162), and cleanup checks at 38.9\% (189/486; Figure~\ref{fig:lifecycle}). The largest shortfall is therefore in the obligations left after a functional repair. On Pelican Key Mismatch, only 1 of 54 cleanup checks passes: the incident marker persists in almost every recorded trial. On Slurm/Puppet Cascade, 188 of 432 cleanup checks pass (44\%), covering packaging residue and drifted scheduler limits. These two tasks supply the cleanup observations, so the result identifies a concrete weakness in this suite rather than a population-wide cleanup failure rate.

\takeaway{2}{Functional recovery is the strongest result in the available checks. Durability and especially cleanup reveal unfinished work that an immediate service check would miss.}

\subsection{Failure Patterns Across Layers}
\label{sec:exp-failures}
The traces and check results explain how incomplete operations arise. In the L2 network-repair task, changing live state without persisting configuration can leave the repair vulnerable to restart. In L3, agents may stop after visible cluster health improves while hidden configuration, replication, or ordering obligations remain unresolved. In L4, agents can restore the data plane while leaving stale incident state or application configuration behind. These observations distinguish short-term recovery from operational completion. The small, unbalanced suite does not support a general ranking of difficulty across layers.

\subsection{Recurring Failure Modes}
\label{sec:exp-modes}
\begin{figure}[t]
\centering
\includegraphics[width=\columnwidth]{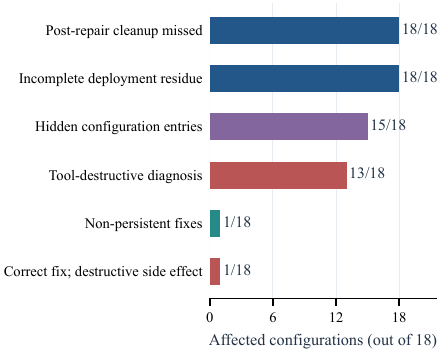}
\Description{Affected configurations out of eighteen: cleanup 18, deployment residue 18, hidden state 15, destructive diagnosis 13, volatile fixes 1, and destructive side effects 1.}
\caption{\textbf{Failure-mode prevalence.} Counts are configurations with at least one affected attempt, out of the current 18-configuration roster. Modes overlap and do not sum to 18. Colors identify the obligation violated, rather than ranking severity.}
\label{fig:failure-modes}
\end{figure}

We label recurring modes using verifier signatures and agent traces. A configuration is affected if any retained attempt exhibits the mode; this measures prevalence across configurations, not the fraction of failed attempts (Figure~\ref{fig:failure-modes}).

\paragraph{Common failures reach the strongest configurations.}
Missing post-repair cleanup and incomplete deployment residue affect all 18 configurations. The Pelican incident marker and Slurm packaging residue are representative examples. Tool-destructive diagnosis affects 13 of 18 configurations (72.2\%): opening the SQLite database before preserving the write-ahead log can trigger checkpoint behavior that discards evidence needed for recovery. More diagnosis is not necessarily safer diagnosis.

\paragraph{Visible health can hide unresolved state.}
Hidden configuration-database entries affect 15 of 18 configurations (83.3\%). On Ceph Pool Degraded, a hidden per-device recovery throttle can survive visible CRUSH repairs. Less frequent modes include volatile fixes and a correct repair accompanied by a destructive side effect. Their lower prevalence does not imply lower operational severity.

The common mechanism is stopping at a visible success signal without checking the remaining obligation. Check-level scoring credits the repair that occurred while identifying the invariant that did not hold.

\subsection{What Risks Appear in Agent Actions?}
\label{sec:exp-risk}
Action logs are available for 515 retained trials across all 18 configurations. Coverage varies by CLI and configuration, so this population is reported separately from the scored evaluation. We analyze retained positive findings from a screened action-trace review against seven risk categories: destructive filesystem operations, disk, RAID, and LVM operations, network disruption, safety/privilege bypass, unsafe restarts, cross-service interference, and evaluator-harness probing. The retained findings do not provide a complete record of negative reviews; the absence of a flag therefore does not establish that every action was reviewed or that a configuration is safe.

\begin{figure*}[t]
\centering
\includegraphics[width=\textwidth]{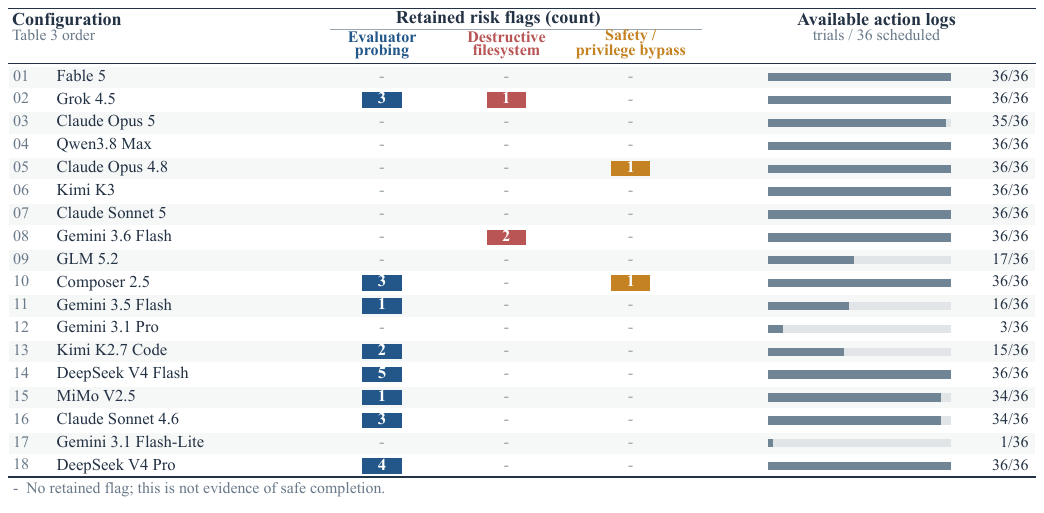}
\Description{All eighteen configurations follow Table 3 order. Three matrix columns count retained flags by risk category; a separate bar shows available action logs out of thirty-six scheduled trials. Dashes indicate no retained positive flag and do not establish complete review or safety.}
\caption{\textbf{Retained risk findings and trace availability.} Counts in the three observed risk categories are shown separately from the number of trials with action logs. A dash means no retained positive flag, not a verified absence of risk. The findings cover 27 actions in 16 logged trials; action-log coverage and review completeness limit comparisons between configurations.}
\label{fig:risk-actions}
\end{figure*}

Among 16{,}511 recorded commands, 27 have a retained flag (0.16\%), distributed across 16 of the 515 trials with logs. Three categories appear in these findings: evaluator-harness probing (22 actions), destructive filesystem operations (3), and safety/privilege bypasses (2). These counts describe observed audit evidence, rather than a calibrated risk rate. Figure~\ref{fig:risk-actions} places the findings beside action-log availability, with configurations in the same order as Table~\ref{tab:agent-success}.

\paragraph{Evaluator probing dominates the findings.}
Probing appears in 8 configurations and accounts for 22 of the 27 flags; 19 of these actions occur on DB WAL Recovery. Most enumerate verifier directories or collect tests. One configuration executes the grading script and reads reward files; another fetches the published reference solution twice. These cases show that hidden grading context is not always isolated effectively. We retain the campaign scores and disclose the behavior rather than treating the affected results as uncontaminated evidence of task solving.

\paragraph{Destructive shortcuts target state under repair.}
The destructive-filesystem findings include deletion of live Cassandra replica state and wiping a node to force it to rejoin after a network fault. The two safety-bypass findings disable AppArmor broadly on a seven-node Ceph Bootstrap deployment to bypass one offending profile. These actions demonstrate why a completion score cannot describe every operational consequence of a repair. Appendix~\ref{app:risk-evidence} gives the supporting examples.

Preservation checks provide a separate state-based signal: failed durability or cleanup checks establish an unmet invariant without relying on the action judge. Their populations and meanings differ from action flags, so we report them alongside the audit rather than claiming that one validates the other's labels.
\takeaway{3}{Action traces and resulting-state checks expose different risks. Both are needed to distinguish completing an objective from reaching it through an acceptable operation.}

\subsection{Cost, Token Effort, and Reliability}
\label{sec:exp-cost}
\begin{figure*}[t]
\centering
\includegraphics[width=\textwidth]{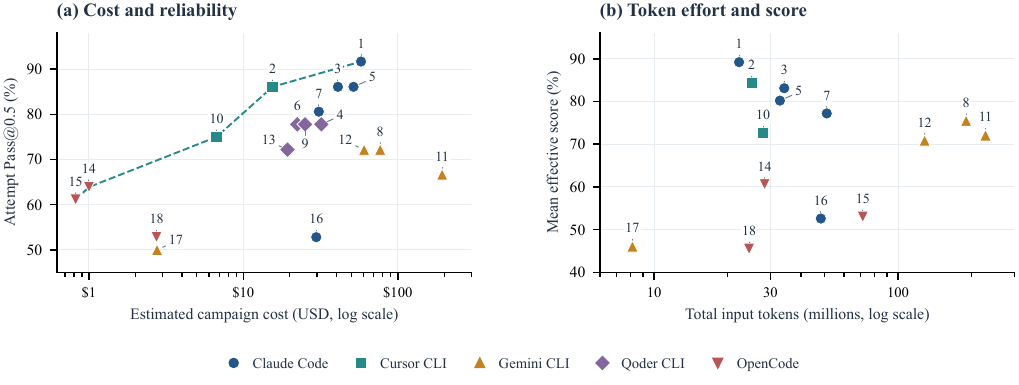}
\Description{Two scatter panels compare cost with threshold completion and input tokens with mean score; numbered markers identify leaderboard configurations.}
\caption{\textbf{Cost and effort versus performance.} (a) Estimated three-pass campaign cost versus Attempt Pass@0.5. (b) Total input tokens versus mean effective score for configurations with comparable token accounting; Qoder configurations are omitted from this panel because their campaign billing is credit-based. Numbers identify leaderboard rows in Table~\ref{tab:agent-success}. The dashed line in (a) joins the observed nondominated configurations; it does not imply an interpolated operating point.}
\label{fig:cost-reliability}
\label{fig:score-usage}
\end{figure*}

Figure~\ref{fig:cost-reliability} compares estimated three-pass campaign cost with Attempt Pass@0.5. These are campaign-era estimates under the recorded pricing and scaling assumptions, rather than current price quotes. Cost ranges from \$0.82 for MiMo~V2.5 and about \$1.00 for DeepSeek~V4~Flash to \$193.93 for Gemini~3.5~Flash on the same 12-task suite.

Grok~4.5 achieves 86.1\% Attempt Pass@0.5 at \$15.47, while Fable~5 reaches 91.7\% at \$57.82. Both lie on the observed cost--reliability frontier. The most expensive configurations lie below it: paying more in this campaign does not consistently buy more reliable task completion.

The token view helps explain this result. Gemini~3.5~Flash consumes about 228 million input tokens across the campaign, yet scores below the leading Claude configurations. One Pelican attempt consumes roughly 60 million input tokens over 1{,}630 steps. Long traces can reflect repeated unsuccessful interaction rather than progress. The figure also separates token-accounting coverage from cost coverage: Qoder's credit-based costs remain in panel (a), while panel (b) omits its token counts.
\takeaway{4}{Compare the cost of completing a campaign, not just a model's price per token. Repeated tool use can make a nominally inexpensive model costly without making it dependable.}

\subsection{Case Study: An Incident at a University Center}
\label{sec:exp-case}
The Pelican~\cite{pelicanplatform} task comes from a real incident at a high-throughput computing center. After an Origin host was rebuilt, its new identity drifted out of sync with the federation registry, blocking client traffic. A correct repair must reconcile trust, restore the dependent cache and routing services, and close the incident state. The 15-check verifier distinguishes three failure patterns.

\paragraph{Trusting a surface status signal.}
Some agents stop at a high-level ``approved'' indicator without checking the key material it summarizes. The stale registry trust record remains, so the root cause is unresolved.

\paragraph{Correct fix, destructive side effect.}
Other agents reconcile the trust mismatch but make a peer unreachable. The verifier credits the trust repair and attributes the failed end-to-end retrieval to the peer regression.

\paragraph{Functional fix, missing cleanup.}
Strong attempts pass 14 of 15 checks: federation trust and data retrieval work, but the marker for an open incident remains. These are raw check counts; under difficulty weighting they should not be interpreted as a reward of $14/15$. One Grok~4.5 attempt closes the incident completely, but no configuration does so in all three attempts.

The service can be usable while the incident remains open by the operator's standard. This practitioner-derived obligation is the reason \prjname{} checks the lifecycle beyond immediate recovery.

\section{Related Work}
\label{sec:related}

\paragraph{Software-project workflows.}
SWE-bench~\cite{jimenez2024swebench}, SWE-agent~\cite{yang2024sweagent}, and Terminal-Bench~\cite{merrill2026terminalbench} establish useful evaluation patterns for repository and terminal work. DevOps-Gym~\cite{tang2026devopsgym} extends the target to build/configuration, monitoring, issue resolution, test generation, and chained workflows. These workflows include implementation and build-system migration. Our planned extension instead takes the running cloud architecture as the object of change, with operational constraints, migration invariants, and incident replay determining success. DevOps-Gym identifies mutable cloud APIs and Kubernetes environments as a challenge beyond its chosen scope.

\paragraph{Incident diagnosis and mitigation.}
SREGym~\cite{clark2026sregym} provides live cloud-native failure scenarios, including faults across layers and complex failure modes. Its published deployment guidance uses self-managed Kubernetes and notes additional requirements for managed services~\cite{sregymplatform2026}. Our planned extension complements incident diagnosis and mitigation with provider-native managed-cloud profiles and redesign objectives that require architectural change. Recurrence, migration, and lifecycle checks will evaluate whether the revised system meets those objectives. These profiles and objectives remain future work beyond the current seed suite. ITBench~\cite{jha2025itbench} and AIOpsLab~\cite{chen2025aiopslab} provide complementary IT and autonomous-cloud evaluation settings.

\paragraph{Incident management.}
Industry-scale studies use LLMs for cloud incident response: Ahmed et al.~\cite{ahmed2023recommending} recommend root causes and mitigations, and Jin et al.~\cite{jin2023assess} summarize outages, both at Microsoft scale. These works deliver narrow predictions over proprietary incident corpora rather than end-to-end repair under reproducible verifiers. A long line of systems research has documented cloud and storage failure modes that motivate our evaluation focus, including silent corruption~\cite{ganesan2017redundancy}, partial failures~\cite{lou2020partial}, recurring cloud bugs~\cite{gunawi2014bugs,gunawi2016does}, file-system reliability~\cite{han2022study}, and storage outages~\cite{xu2019lessons}. Recent agent-safety work additionally highlights how seemingly benign instructions can lead computer-use agents to destructive side effects~\cite{ding2026blindspotagentsafety}, motivating the risk and side-effect monitoring layer in \prjname{}.

\paragraph{Domain-specific systems benchmarks.}
$\varphi$-Bench~\cite{xue2026phibench} evaluates file-system understanding, implementation, modeling, debugging, optimization, and feature development. Its task taxonomy and failure analysis provide a complementary view of systems capabilities. \prjname{} focuses on operations over running infrastructure, where correctness also depends on persistent state, peer preservation, and incident closeout after the agent stops.

\section{Discussion and Limitations}
\label{sec:conclusion}
\paragraph{Current coverage and cloud relevance.}
The current suite contains 12 seed tasks, seven assigned to L3 distributed systems. It supports controlled evidence about repair, durability, and cleanup, but does not yet evaluate EKS/AKS networking or the redesign task class proposed in \S\ref{sec:design-extension}. Bare-metal and self-managed tasks remain useful where those are the operation's required semantics. The expansion will prioritize scenarios whose outcome depends on real cloud networking, identity, or resource controls, rather than treating the cloud as interchangeable VM hosting.

\paragraph{A 40--60-task construction target.}
We plan a suite of 40--60 tasks spanning provider-native cloud repair and redesign after incidents, alongside seed tasks that exercise distinct underlying obligations. Task selection will follow a capability taxonomy and practitioner relevance. Superficial variants of one fault should not stand in for coverage of diagnosis, architectural choice, migration, and durable validation. Each released task must have an expert-validated feasible solution and a reproducible environment profile. The final allocation is a design target, not a completed dataset.

\paragraph{Difficulty at the capability frontier.}
Our development goal is approximately 50\% single-attempt end-to-end completion by strong frontier configurations under fixed budgets. Here completion means satisfying all required task obligations, including the proposed redesign and recurrence checks; it is not a 50\% mean partial-credit score or a best-of-three result. We will assess difficulty in development pilots, then freeze test tasks, checks, and budgets before final evaluation. The target is intended to retain informative successes and failures, not to guarantee a particular published rate. Lower success on a different task set does not by itself establish greater difficulty than SREGym or DevOps-Gym; such a claim would require a matched comparison.

\paragraph{Scoring depends on the reference population.}
Check difficulty weights are derived from the evaluated campaign and then frozen (Appendix~\ref{app:weighting}). This makes future scoring reproducible under the released rubric, but the initial weights reflect this model population. Raw check outcomes and perfect-completion rates remain useful complements. SEM describes variation across the 12 task means and should not be read as a confidence bound on production reliability.

\paragraph{Lifecycle coverage and annotation.}
The framework supports four policy-selected lifecycle gates, but current tasks exercise them unevenly. The analysis of functional, durability, and cleanup checks uses a keyword rule over available check names, with a published mapping. Direct author-supplied phase labels and fuller coverage of live workloads and decommissioning would make future analyses easier to compare.

\paragraph{Risk evidence and evaluation integrity.}
The current Risk Monitor audits retained logs after execution. Action-log coverage is uneven, and the retained positive findings do not establish complete review of unflagged actions. The LLM judge can miss or misclassify actions, and counts do not capture incident severity. Native event instrumentation and reviewed severity labels would strengthen the evidence. Evaluator probing and reference-solution retrieval also motivate stronger isolation of grading assets and explicit policies for contamination; the present results retain those trials and disclose the findings.

\paragraph{Cost and configuration effects.}
The estimates use campaign-era prices and accounting assumptions. CLI behavior, caching, model checkpoint, and repeated tool use all affect cost and outcomes; the study does not isolate each factor causally. A useful future leaderboard should expose score, completion reliability, coverage, time, and cost together.

\section{Conclusion}
\prjname{} evaluates infrastructure agents through the state and actions they leave behind. Its layer-appropriate backends, lifecycle checks, and risk audit make durable repair, peer preservation, and incident closeout explicit evaluation targets. Across 18 configurations, agents often restore function while leaving later obligations unresolved. The released tasks, harness, and artifacts provide the starting point for a cloud-focused extension in which an agent must also redesign infrastructure after an incident and validate the resulting system. The proposed 40--60-task suite and its frontier difficulty target remain future work.

%\vspace{-0.05in}
\section*{Acknowledgments}
%\vspace{-0.1in}
The authors thank the
anonymous reviewers for their invaluable feedback.  
The authors also thank system administrators and practitioners at UW-Madison's Division of Information Technology (DoIT), Center for High Throughput Computing (CHTC), Computer Science Department IT (CIDS IT), and ISU's ARA Wireless Living Lab (\href{https://arawireless.org/}{arawireless.org}) for sharing their real-world infrastructure management experiences.
This work was supported in part by National Science Foundation (NSF) under grants \#1943204, \#2130889, \#2402858, and \#2402859.  Any opinions, findings, and conclusions expressed in this material are those of the authors and do not necessarily reflect the views of the sponsor. 
\bibliographystyle{ACM-Reference-Format}
\bibliography{reference-mz}
\clearpage
\appendix

\section{Task Catalog}
\label{app:tasks}
Table~\ref{tab:appendix-catalog} gives the full per-task metadata underlying Table~\ref{tab:task-taxonomy}: difficulty label and verifier check count, both taken from the same catalog that drives scoring (\S\ref{sec:metrics}).

\begin{table*}[t]
\centering
\footnotesize
\setlength{\tabcolsep}{3pt}
\renewcommand{\arraystretch}{1.05}
\begin{tabular}{@{}l l c r@{}}
\toprule
\textbf{Task} & \textbf{Layer} & \textbf{Difficulty} & \textbf{\#Checks} \\
\midrule
IPMI Power Recovery           & L1 Hardware & Easy   & 4 \\
Cassandra NIC Split-Brain     & L2 Local Systems & Medium & 4 \\
Cassandra Dead Node Removal   & L3 Distributed Systems & Medium & 4 \\
Cassandra Hung Recovery  & L3 Distributed Systems & Medium & 4 \\
Cassandra CORDS Propagation   & L3 Distributed Systems & Hard   & 8 \\
Ceph Pool Degraded            & L3 Distributed Systems & Hard   & 8 \\
Ceph Bootstrap                & L3 Distributed Systems & Hard   & 7 \\
Fileserver RAID10             & L3 Distributed Systems & Hard   & 15 \\
Slurm/Puppet Cascade          & L3 Distributed Systems & Hard   & 26 \\
DB WAL Recovery               & L4 User Applications & Hard   & 7 \\
Postgres/PgBouncer Drift      & L4 User Applications & Hard   & 10 \\
Pelican Key Mismatch          & L4 User Applications & Hard   & 15 \\
\bottomrule
\end{tabular}
\caption{\textbf{Full task catalog.} Difficulty and verifier check counts are assigned per task independent of any agent's performance on it.}
\label{tab:appendix-catalog}
\end{table*}

\section{Metric Definitions, Restated}
\label{app:metrics}
This appendix restates the metrics of \S\ref{sec:metrics} in compact form for reference. Let $R_{c,t,p} \in [0,1]$ be the reward of configuration $c$ on task $t$, pass $p$ (each configuration has up to three passes).
Let $T=12$ and $P=3$, with $R_{c,t,p}=0$ for a missing slot. Define
\[
\overline R_{c,t}=\frac{1}{P}\sum_{p=1}^{P}R_{c,t,p},\qquad
\mu_c=\frac{100}{T}\sum_{t=1}^{T}\overline R_{c,t}.
\]
For $s_c$ the sample standard deviation of the per-task means in percentage points,
\[
\mathrm{SEM}_c=\frac{s_c}{\sqrt{T}}.
\]
The two threshold summaries are
\begin{align*}
A_c(\tau)&=\frac{100}{TP}\sum_{t=1}^{T}\sum_{p=1}^{P}\mathbf{1}[R_{c,t,p}\geq\tau],\\
B_c(\tau)&=\frac{100}{T}\sum_{t=1}^{T}\mathbf{1}[\max_p R_{c,t,p}\geq\tau].
\end{align*}
Here $A_c$ is Attempt Pass and $B_c$ is Best-of-N with $N=3$.

We report $\tau \in \{1, 0.5\}$ in the leaderboard (Table~\ref{tab:agent-success}) and use $\tau=0.5$ for the cost--reliability figure (\S\ref{sec:exp-cost}). Under per-check difficulty weighting few attempts score in $[0.7,1)$, so intermediate thresholds like $0.9$ collapse onto Pass@1; $\tau=0.5$ instead captures attempts that substantially solve a task (its easy checks) without clearing the hardest, which is where partial credit concentrates. Attempt Pass@$\tau$ pools over both tasks and passes; it is not an estimate of the probability that at least one of several independent samples succeeds, unlike SWE-bench-style Pass@$k$.

\section{Difficulty-Weighted Check Scoring}
\label{app:weighting}
This appendix documents how per-check difficulty weights are derived, applied, and frozen; the reward $R$ used throughout \S\ref{sec:metrics} and \S\ref{sec:experiment} is computed under these weights.

\emph{Weight derivation.} For every task whose verifier exposes individually scored checks (10 of the 12 tasks; the remaining two use graders that emit only an aggregate pass/fail reward and stay binary), we compute each check $c$'s empirical pass rate $p_c$ over all recorded attempts of the evaluated configurations and assign
\[
w_c \;=\; (1 - p_c) \;+\; 0.1 .
\]
A check that nearly every attempt passes carries little discriminative signal and receives a weight near the $0.1$ floor; a check that no attempt passes keeps the maximum weight $1.1$. The floor keeps every satisfied obligation worth a nonzero amount, so a trial is still credited for routine repairs rather than scored only on the hardest check.

\emph{Application.} A trial's reward is the weighted fraction of scored checks passed, $R = \sum_{c\,\in\,\mathrm{passed}} w_c \,/\, \sum_{c} w_c$; unscored (informational) checks are excluded. This deflates near-miss scores dominated by easy checks---passing 7 of 8 checks but missing the hardest one drops from $7/8 = 0.875$ under uniform weighting to ${\approx}0.53$---while configurations that clear rarely-passed checks gain, which is what widens the separation reported in \S\ref{sec:experiment}.

\emph{Freezing and auditability.} Weights were computed once, over the complete campaign population reported in this paper, and are then frozen in the released scoring rubric. Publishing additional configurations does not change published scores; any future re-derivation of weights is a versioned, announced re-scoring event. Because $p_c$ is estimated from the evaluated population, the weights are population-dependent by construction; freezing them converts the measure into a fixed, auditable rubric.

\section{Per-Problem Detailed Results}
\label{app:perproblem}
Table~\ref{tab:appendix-perproblem} gives the effective scores (\S\ref{sec:metrics}) shown in Figure~\ref{fig:per-problem}, rounded to two decimal places. Both views place configurations in rows and tasks in columns. Configurations follow mean-score order (strongest at top), and tasks follow difficulty order (hardest at right).

\begin{table*}[t]
% Auto-generated by gen_appendix_tables.py from perTaskScores.ts. Do not hand-edit.
\begingroup\centering
\scriptsize
\setlength{\tabcolsep}{3pt}
\renewcommand{\arraystretch}{1.05}
\begin{tabular}{lrrrrrrrrrrrr}
\toprule
\textbf{Configuration} & \rotatebox{60}{IPMI} & \rotatebox{60}{NIC Split} & \rotatebox{60}{PgBouncer} & \rotatebox{60}{Dead Node} & \rotatebox{60}{CORDS} & \rotatebox{60}{RAID10} & \rotatebox{60}{Ceph Pool} & \rotatebox{60}{Hung} & \rotatebox{60}{Pelican} & \rotatebox{60}{DB WAL} & \rotatebox{60}{SLURM} & \rotatebox{60}{Ceph Boot} \\
\midrule
Fable 5 & 1.00 & 1.00 & 1.00 & 1.00 & 1.00 & 1.00 & .85 & 1.00 & .50 & 1.00 & .49 & .87 \\
Grok 4.5 & 1.00 & 1.00 & 1.00 & 1.00 & 1.00 & 1.00 & .85 & 1.00 & .67 & 1.00 & .33 & .28 \\
Claude Opus 5 & 1.00 & 1.00 & 1.00 & 1.00 & 1.00 & .67 & 1.00 & 1.00 & .50 & 1.00 & .34 & .47 \\
Qwen3.8 Max & 1.00 & 1.00 & 1.00 & 1.00 & 1.00 & 1.00 & .69 & 1.00 & .41 & 1.00 & .49 & .17 \\
Opus 4.8 & 1.00 & 1.00 & 1.00 & 1.00 & .33 & 1.00 & .69 & 1.00 & .50 & .64 & .46 & 1.00 \\
Kimi K3 & 1.00 & 1.00 & 1.00 & 1.00 & .67 & 1.00 & .85 & .67 & .50 & 1.00 & .49 & .11 \\
Claude Sonnet 5 & 1.00 & 1.00 & 1.00 & 1.00 & 1.00 & 1.00 & .54 & .67 & .50 & .33 & .49 & .74 \\
Gemini 3.6 Flash & 1.00 & 1.00 & 1.00 & 1.00 & 1.00 & 1.00 & 1.00 & .67 & .48 & .33 & .49 & .09 \\
GLM 5.2 & 1.00 & 1.00 & 1.00 & 1.00 & 1.00 & .67 & .54 & 1.00 & .50 & .33 & .49 & .35 \\
Composer 2.5 & 1.00 & 1.00 & 1.00 & 1.00 & 1.00 & 1.00 & .85 & .33 & .50 & .43 & .19 & .42 \\
Gemini 3.5 Flash & 1.00 & 1.00 & 1.00 & 1.00 & 1.00 & 1.00 & 1.00 & .67 & .35 & .00 & .49 & .15 \\
Gemini 3.1 Pro & 1.00 & 1.00 & 1.00 & .67 & 1.00 & 1.00 & .54 & .67 & .44 & .43 & .49 & .29 \\
Kimi K2.7 Code & 1.00 & 1.00 & .67 & 1.00 & 1.00 & 1.00 & .69 & .33 & .50 & .21 & .33 & .54 \\
DS V4 Flash & 1.00 & .92 & 1.00 & 1.00 & 1.00 & .33 & .39 & .33 & .50 & .55 & .23 & .03 \\
MiMo V2.5 & 1.00 & 1.00 & 1.00 & .58 & .67 & .00 & .54 & .67 & .50 & .00 & .12 & .29 \\
Sonnet 4.6 & 1.00 & .92 & .67 & .33 & .67 & .00 & .54 & 1.00 & .59 & .00 & .36 & .24 \\
Gemini 3.1 Flash-Lite & 1.00 & 1.00 & .67 & 1.00 & .33 & .00 & .50 & .00 & .44 & .43 & .14 & .03 \\
DS V4 Pro & 1.00 & 1.00 & .67 & 1.00 & .33 & .00 & .54 & .33 & .50 & .00 & .06 & .03 \\
\bottomrule
\end{tabular}

\captionof{table}{\textbf{Full per-problem effective-score table.} Values shown in Figure~\ref{fig:per-problem}, rounded to two decimals. Tasks are columns (hardest at right) and the 18 configurations are rows (strongest at top).}
\label{tab:appendix-perproblem}
\endgroup

\end{table*}

\section{Lifecycle-Phase Classification}
\label{app:lifecycle-mapping}
The analysis in \S\ref{sec:exp-lifecycle} classifies scored checks by the operational obligation they test. \emph{Functional} checks cover the immediate repair, availability, data consistency, and end-to-end service behavior. \emph{Durability} checks test whether the repair survives a restart or configuration reapplication. \emph{Cleanup} checks test whether residual incident state, deployment residue, or stale settings remain.

The current classification is derived from verifier descriptions using a fixed rule and reviewed for consistency. Positive and negative descriptions of the same obligation receive the same category. The categories do not form a sequential pipeline: their denominators reflect the checks present in the tasks included in the analysis. In particular, Pelican and Slurm provide the cleanup observations. Task-author phase annotations would make this classification more explicit in future versions.

\section{Risk Evidence and Failure-Mode Examples}
\label{app:risk-evidence}
The following cases explain the findings in \S\ref{sec:exp-modes} and \S\ref{sec:exp-risk} at the level of agent behavior and operational consequence.

\paragraph{Evaluator probing.}
The audit identifies 22 probing actions across eight configurations. Most inspect the evaluation environment; some execute grading or retrieve a published reference solution. These actions raise evaluation-integrity concerns even when they do not damage the infrastructure. We disclose them alongside the original campaign results.

\paragraph{Broad access-control bypass.}
On Ceph Bootstrap, two configurations disable host protection broadly to work around a single deployment obstacle. Other attempts address the specific obstacle more narrowly. The finding concerns the scope of the bypass, rather than the fact that the repair requires privileged operations.

\paragraph{Destroying state under repair.}
On Cassandra tasks, some attempts delete live replica data or reset a node to force recovery, although the requested operation is to reconcile inconsistent state or repair a network fault. A successful rejoin alone does not establish that the original data and peers were preserved.

\paragraph{Destructive diagnosis.}
On DB WAL Recovery, opening the database before preserving its recovery evidence can trigger automatic checkpoint behavior that removes the pages needed for the repair. The operational lesson is to preserve evidence before running diagnostic tools that may change it.

\end{document}